\documentclass[preprint,12pt,authoryear]{elsarticle}

\usepackage{natbib}
\usepackage{multirow}
\usepackage{amsmath}
\usepackage{amssymb}
\usepackage{graphicx}
\usepackage{hyperref}
\usepackage{subfigure}
\usepackage{soulutf8}
\usepackage{natbib}
\usepackage[title]{appendix} 
\usepackage{lineno}
\usepackage{placeins}
\usepackage{graphicx} 

\usepackage{xcolor}
\sethlcolor{yellow}
\usepackage{geometry}

\soulregister\cite7
\soulregister\ref7
\soulregister\citep7

\journal{Transportation Research Part C}

\begin{document}

\begin{frontmatter}



\title{A Foundational Individual Mobility Prediction Model based on Open-Source Large Language Models} 


\author[label1]{Zhenlin Qin} 
\author[label2]{Leizhen Wang} 
\author[label1]{Yancheng Ling} 
\author[label3]{Francisco Câmara Pereira} 
\author[label1]{Zhenliang Ma\corref{cor1}} 

\cortext[cor1]{Corresponding author.}

\affiliation[label1]{organization={Department of Civil and Architectural Engineering, KTH Royal Institute of Technology},
            country={Sweden}}

\affiliation[label2]{organization={Department of Data Science and Artificial Intelligence, Monash University},
            country={Australia}}

\affiliation[label3]{organization={Department of Technology, Management and Economics Intelligent Transportation Systems, Technical University of Denmark},
            country={Denmark}}


\begin{abstract}
Individual mobility prediction plays a key role in urban transport, enabling personalized service recommendations and effective travel management. It is widely modeled by data-driven methods such as machine learning, deep learning, as well as classical econometric methods to capture key features of mobility patterns. However, such methods are hindered in promoting further transferability and robustness due to limited capacity to learn mobility patterns from different data sources, predict in out-of-distribution settings (a.k.a ``zero-shot"). To address this challenge, this paper introduces MoBLLM, a foundational model for individual mobility prediction that aims to learn a shared and transferable representation of mobility behavior across heterogeneous data sources. Based on a lightweight open-source large language model (LLM), MoBLLM employs Parameter-Efficient Fine-Tuning (PEFT) techniques to create a cost-effective training pipeline, avoiding the need for large-scale GPU clusters while maintaining strong performance. We conduct extensive experiments on six real-world mobility datasets to evaluate its accuracy, robustness, and transferability across varying temporal scales (years), spatial contexts (cities), and situational conditions (e.g., disruptions and interventions). MoBLLM achieves the best F1 score and accuracy across all datasets compared with state-of-the-art deep learning models and shows better transferability and cost efficiency than commercial LLMs. Further experiments reveal its robustness under network changes, policy interventions, special events, and incidents. These results indicate that MoBLLM provides a generalizable modeling foundation for individual mobility behavior, enabling more reliable and adaptive personalized information services for transportation management.

\end{abstract}


\end{frontmatter}



\section{Introduction}
Individual mobility prediction studies human mobility patterns at an individual level to forecast future movements. It facilitates a broad spectrum of applications in transportation domain, such as personalized services of POI or route recommendations and incentive mobility management \citep{ma2022individual}. Data-driven methods are dominant in this field including conventional Markov-based machine learning models \citep{Gambs2012, mathew2012, Lu2013} and deep learning (DL) models such as recurrent neural network (RNN) or Transformer \citep{Feng2018, kong2018hst, xue2021mobtcast}. However, they are still limited to specialized training on the mobility dataset from a single city or data source, leading to low transferability of adapting to different travel contexts in different cities. It would be valuable to develop a foundational individual mobility prediction model capable of addressing these challenges.

Recently, Large Language Models (LLMs) have made significant progress, far surpassing previous state-of-the-art models such as BERT \citep{devlin2019bert} and T5 \citep{raffel2020exploring}. The LLMs become the new foundational model in Natural Language Processing (NLP) domain. In Artificial Intelligence (AI) systems, a foundational model is any model that is trained on broad data (generally using self-supervision at scale) that can be adapted (e.g., fine-tuned) to a wide range of downstream tasks \citep{bommasani2021opportunities}. As fine-tuning techniques advance, LLMs can be applied to developing foundational models for other various disciplines \citep{wang2025discovery, awais2025foundation}.

LLMs have also inspired a range of applications in urban mobility \citep{ma2025llm, wang2024ai}, particularly in mobility prediction tasks. Specifically, typical LLM based methods like In-Context Learning (ICL) \citep{brown2020language} and Chain-of-Thought (CoT) \citep{Wang_2023_psp, wei2022chain} enable modeling mobility prediction problems with natural language \citep{Wang_2023, liang2024exploring, feng2024agentmove}. However, hallucination issues may arise when applying these methods, as their use of commercial LLMs lacks specialized training on mobility datasets. 

The hallucination is first defined in natural language generation as generated content that is either nonsensical or unfaithful to the provided source content \citep{huang2025survey}, which can be similarly defined as unexpected LLMs' outputs under mobility prediction context. Therefore, the existing prompt engineering based methods are limited in generalizing prediction under different contexts (lack foundational capability). Besides, fine-tuning based on pretrained LLMs are applied to developing the individual mobility prediction framework \citep{li2024limp, gong2024mobility}. They begin by aligning visited locations with specific intents, then leverage LLMs to predict the next location based on this enriched semantic context. From model validation perspectives, most prior studies focus primarily on prediction / accuracy but have not sufficiently demonstrated the LLMs' capabilities of transferability and robustness across different data sources and prediction scenarios.

Inspired by the notion of foundation models in artificial intelligence, we use the term foundational model in a transportation-science sense rather than in an architecture-centric AI sense. In this paper, a foundational mobility model refers to a reusable, general-purpose behavioral representation that supports multiple mobility-related prediction and policy-relevant tasks across heterogeneous cities, time periods, and contextual conditions.

Unlike traditional transportation models that are calibrated separately for each dataset, city, or policy scenario, a foundational mobility model aims to learn a shared representation of individual mobility behavior that can be transferred, adapted, and reused across contexts. The novelty lies not in proposing a new neural architecture, but in establishing a unified modeling substrate for mobility prediction that remains valid across spatial scales (cities), temporal scales (years), and situational conditions (e.g., disruptions, pricing, and interventions).

Under this definition, LLMs serve as a practical implementation mechanism for building such a foundation, because they provide a flexible representation of heterogeneous mobility data, traveler attributes, and contextual factors within a single transferable model. The key contributions include:

\begin{itemize}
\item We introduce a \emph{general-purpose mobility foundation model} for individual mobility prediction that provides a unified, transferable behavioral representation across heterogeneous data sources, cities, temporal periods, and situational contexts, enabling consistent modeling of mobility behavior beyond case-specific calibration.

\item We develop a scalable instruction-based adaptation framework that enables large language models to be efficiently specialized for diverse mobility prediction tasks and user-specific data formats, supporting flexible transfer across datasets without retraining from scratch.

\item We conduct extensive empirical evaluation on multi-city mobility datasets—including GPS trajectories, Automated Fare Collection (AFC) records, and check-in data—to systematically assess prediction accuracy, cross-city transferability, temporal generalization, and robustness to contextual variation, thereby establishing both the strengths and empirical boundaries of the proposed foundation model paradigm.

\end{itemize}







The remainder of the paper is organized as follows: Section \ref{sec2} reviews the recent approaches related to modeling individual mobility prediction in the literature. Section \ref{sec3} defines the studied problem and proposes the framework, including instruction data generation and fine-tuning based on open-source LLMs. Case studies are conducted on real-world datasets are presented in Section \ref{sec4}. The final section concludes the main findings and discusses future work.

\section{Related Work}
\label{sec2}
The problem of individual mobility prediction consists of predicting an individual’s possible next location or trip information given their historical mobility sequence data \citep{ma2022individual}. In early stage, the Markov based methods are widely applied to modeling location transition probabilities, i.e, the next location prediction is based on several previous visited locations \citep{Gambs2012, mathew2012, Lu2013}. Although these methods are intuitive and interpretable, they have low performance due to the difficulty in learning complex mobility patterns. Therefore, we focus on literature based on deep learning and LLM.

\subsection{Deep Learning based Methods}
Deep learning based methods commonly model individual mobility prediction by capturing the spatiotemporal features of an individual's historical trip sequence. For example, \citet{Liu2016} models local temporal and spatial contexts by extending an recurrent neural network (RNN) layer with time-specific and distance-specific transition matrices. DeepMove \citep{Feng2018} adopted a sequence to sequence (seq2seq)-based framework to capture the multi-level periodicity for mobility prediction from lengthy and sparse trajectories. The DeepTrip model \citep{zhang2023} extended the DeepMove to predict the next trip information of the AFC metro data. Other RNN-based models adopted the Long Short-Term Memory (LSTM) architecture to model the prediction problem \citep{kong2018hst, li2020hierarchical, Chen2020}. 

Besides, several Transformer-based models show stronger capacities in capturing spatiotemporal context features of human trajectories and faster inference speed than the RNN-based models. Specifically, \citet{xue2021mobtcast} used a Transformer-based framework to learn temporal, semantic, social, and geographical contexts from the history of place sequences to predict the next place. \citet{hong2023context} adopted a multi-head self-attention mechanism model to infer the context of historical location visits, visit time, activity duration, and POI to infer individuals' next location. 

The above methods based on RNN or Transformer have achieved significantly better performance than the classical machine learning models like Markov but are still difficult to address the issue of low transferability. These methods can only be generalized to the samples aligned with the distribution of their training data since their relatively small parameter sizes limit the capability of learning the representations of mobility patterns from different data sources. 

\subsection{Large Language Model based Methods}
Common LLM-based methods for mobility prediction employ In-Context Learning (ICL) approaches to prompt LLMs with prediction task descriptions and historical mobility data, as well as Chain-of-Thought (CoT) to guide LLMs to capture complex semantic information. Such approaches enable LLM to make predictions in a zero-shot way without requiring additional training or gradient update. For example, LLM-Mob \citep{Wang_2023} adopted ICL to realize zero-shot individual mobility prediction. Similarly, LLM-MPE \citep{liang2024exploring} predicted human mobility under public events by leveraging the unprecedented ability of LLM to process textual data. LingoTrip \citep{qin2025lingotrip} prompted LLMs to capture trip purposes, choice preferences and choice variability from individual's historical trip data to comprehensively infer the next trip location. Besides, agentic LLM frameworks integrating ICL, CoT and Retrieval-Augmented Generation (RAG) are applied to extending mobility prediction with more dimensions such as activity or intent inference. Specifically, AgentMove \citep{feng2024agentmove} achieved a generalized prediction of mobility for any city worldwide by introducing POI data from cities to capture the knowledge of the world. ZHMF \citep{li2025zero} introduced a framework that integrates short-term trajectory records with structured long-term memory retrieved by RAG technology. LLMob \citep{wang2024large} adopted an agentic LLM framework that accounts for individual activity patterns and motivations. While these methods exhibit promising application potential, the performance still has significant room for improvement since they use commercial LLMs trained on millions of general corpora without training on any mobility dataset. Meanwhile, the commercial LLMs lacking specialized knowledge may cause hallucination issues when answering domain-specific questions \citep{huang2025survey}. 

To solve these problems, fine-tuning methods based on open-source pre-trained LLMs such as Llama \citep{touvron2023llama} adapt the LLMs to individual mobility datasets. Specifically, LIMP \citep{li2024limp} proposed a mixed framework that uses GPT-4o to generate high-quality intent data for training a local open-source LLM. Mobility-LLM \citep{gong2024mobility} utilized LLMs to infer travel preferences using historical mobility data and POI information. Although these methods enable LLMs to capture the mobility intents or preferences of individual by fine-tuning, they have not demonstrated the capability of adapting to different data sources across multiple cities and diverse prediction scenarios such as disruptions and interventions. 

The reviewed mobility-related works based on LLMs are summarized in Table \ref{table_method_comp} according to the primary modeling methods, data and problem settings. Overall, the existing methods based on prompt engineering or fine-tuning show high potential in modeling the individual mobility prediction problem and yield a wide range of valuable applications. However, they are still insufficient to build a foundational model for individual mobility prediction due to the unsolved issues in adapting to diverse prediction tasks across varying spatiotemporal scales and conditional scenarios.

\begin{table}[]
\centering
\scriptsize
\caption{Comparison between the LLM-based related Methods and MoBLLM.}
\begin{tabular}{lllll}
\hline
Model        & Main Method                                                        & Data Type                                                                       & Foundational Capability & Task                                                                                   \\ \hline
LLM-Mob      & ICL and CoT                                                        & Check-in, GPS trajectory                                                        & Limited in accuracy                                                & Individual mobility prediction                                                         \\
LLM-MPE      & ICL and CoT                                                        & Taxi ridership                                                                  & Limited in accuracy                                                & \begin{tabular}[c]{@{}l@{}}Mobility prediction under public \\ events\end{tabular}     \\
LingoTrip    & ICL and CoT                                                        & Metro trip                                                                      & Limited in accuracy                                                & Individual mobility prediction                                                         \\
LLMob        & Agentic LLM                                                        & Check-in                                                                        & Limited in accuracy                                                & Trajectory generation                                                                  \\
ZHMF         & Agentic LLM                                                        & Check-in                                                                        & Limited in accuracy                                                & \begin{tabular}[c]{@{}l@{}}Intent-aware, cold-start mobility\\ prediction\end{tabular} \\
AgentMove    & Agentic LLM                                                        & Check-in, GPS trajectory                                                        & Limited in accuracy                                                & Individual mobility prediction                                                         \\
LIMP         & \begin{tabular}[c]{@{}l@{}}Agentic LLM,\\ fine-tuning\end{tabular} & Call detail records                                                             & Limited in data scope                                             & Intent-aware mobility prediction                                                       \\
Mobility-LLM & \begin{tabular}[c]{@{}l@{}}ICL and CoT,\\ fine-tuning\end{tabular} & Check-in                                                                        & Limited in data scope                                             & Intent-aware mobility prediction                                                       \\
MoBLLM       & \begin{tabular}[c]{@{}l@{}}ICL and CoT,\\ fine-tuning\end{tabular} & \begin{tabular}[c]{@{}l@{}}Check-in, GPS trajectory, \\ metro trip\end{tabular} & Strong                                                             & Individual mobility prediction                                                         \\ \hline
\end{tabular}
\label{table_method_comp}
\end{table}

\section{Methodology}
\label{sec3}
\subsection{Problem Definition}
\label{sec_task}
Most mobility prediction studies focus on predicting the next location given a series of historical mobility records of the individual \citep{ma2022individual}. Considering GPS trajectory data, check-in data, and AFC data used in these studies, we define four base tasks of individual mobility prediction tailored to different types of mobility data.

\emph{Task 1 (GPS trajectory location prediction):} Given a user's sequence of $n$ chronological stays: $\textbf{S} = (S_1, S_2, \dots, S_n)$, the task is to predict the next location of $S_{n+1}$. A stay $S_i$ includes the start time of stay $t_i$, day of the week $w_i$, duration $dur_i$ of $S_i$, and location $l_i$. Note that the location $l_i$ denotes the corresponding GPS point in the original trajectory after discretization and the duration is $dur_i=t_i-t_{i-1}$.

\emph{Task 2 (check-in location prediction):} Given a user's sequence of $n$ chronological stays: $\textbf{S} = (S_1, S_2, \dots, S_n)$, the task is to predict the next location of $S_{n+1}$. A stay $S_i$ includes the start time of stay $t_i$, day of the week $w_i$, and location $l_i$ (physical location / point of interest).

\emph{Task 3 (next trip origin prediction):} Given a user's sequence of $n$ chronological trip activities: $\textbf{S} = (S_1, S_2, \dots, S_n)$, the task is to predict the next trip end location of $S_{n+1}$. A trip activity $S_i$ includes the start time $t_{s_i}$, day of the week $w_i$, duration $dur_i$ of $S_i$, start location $l_{s_i}$, and end location $l_{e_i}$. We define end time $t_{e_i}$, and then the trip activity duration is $dur_i=t_{e_i}-l_{s_i}$.

\emph{Task 4 (next trip destination prediction):} Given a user's sequence of $n-1$ chronological trips: $\textbf{Tr} = (Tr_1, Tr_2, \dots, Tr_{n-1})$, the task is to predict the next destination station of $Tr_n$. A trip $Tr_i$ includes start time of trip $t_{o_i}$, day of the week $w_i$, origin station $s_{o_i}$, duration $dur_i$ of last stay occurring before $Tr_i$, and destination station $d_i$. We define end time of trip $t_{d_i}$, and then the stay duration is $dur_i=t_{o_i} - t_{d_{i-1}}$.

\begin{figure}[htbp]
\centering
\subfigure[]{
\begin{minipage}{0.7\linewidth}
\centering
\includegraphics[width=1\textwidth]{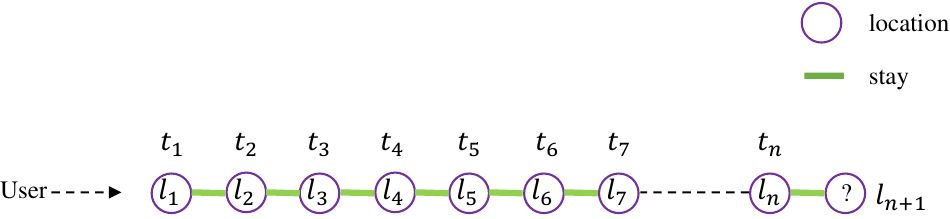}
\end{minipage}
\label{eg_fig1-1}
}
\subfigure[]{
\begin{minipage}{0.7\linewidth}
\centering
\includegraphics[width=1\textwidth]{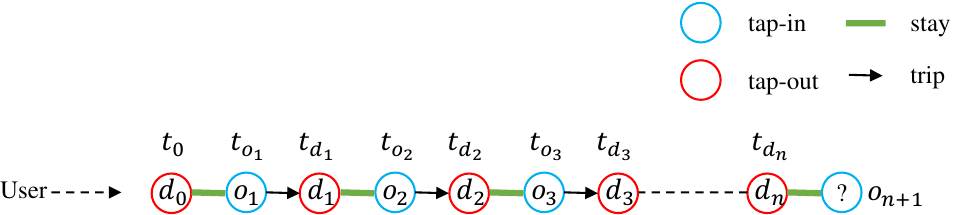}
\end{minipage}
\label{eg_fig1-2}
}

\subfigure[]{
\begin{minipage}{0.7\linewidth}
\centering
\includegraphics[width=1\textwidth]{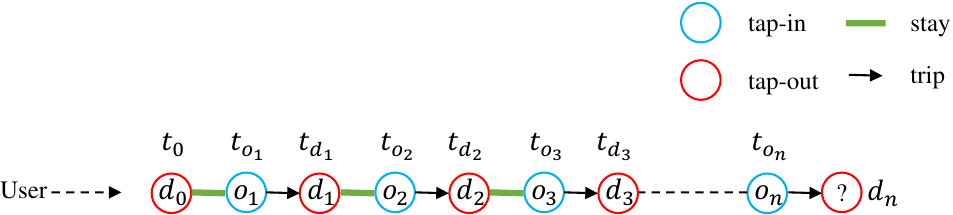}
\end{minipage}
\label{eg_fig1-3}
}
\centering
 \caption{The examples of base tasks. The subfigure (a) presents an example of Task 1 or Task 2, and Tasks 3 and 4 are respectively illustrated by subfigures (b) and (c).}
\end{figure}

The definitions of Tasks 1 and 2 are commonly used in studies conducted on GPS trajectory data or check-in data \citep{Wang_2023, li2024limp, feng2024agentmove} as shown in Figure \ref{eg_fig1-1}. The GPS trajectory data are transformed into discrete representations due to the sparsity of their original coordinate format, making it difficult to model the semantic features of a location. For certain check-in data using the definition of Task 2, the duration information between two adjacent stays can be omitted due to a probable long time span, providing little effective activity information. 

Tasks 3 and 4 are generally defined in the studies conducted on AFC trip data. Here, we adopt an activity-based prediction framework assuming that a stay between two trips contains only one activity, as shown in Figures \ref{eg_fig1-2} and \ref{eg_fig1-3}, which may benefit the model performance \citep{mo2021individual, qin2025lingotrip}. Specifically, we define $d_0=home$ and $t_0=4:00AM$, which means the content of the first activity can be confirmed as staying home. The activity after the last trip of the day is not considered because it is exactly staying home the next day. The information of a trip mainly includes origin station $o$, destination station $d$, start time of trip $t_{o}$, end time of trip $t_{d}$, week of day $w$, which are commonly employed in trip prediction studies \citep{zhang2023deeptrip, zhao2018individual}. However, we extend a trip with the duration of the last stay defined in Task 4, supposing it is associated with certain activity information. Given the context of prediction on AFC trip data, we can align some notations of Task 3 with the trip definition where $t_{s_i}=t_{d_i}$, $t_{e_i}=t_{o_{i+1}}$, $l_{s_i}=d_i$ and $l_{e_i}=o_{i+1}$.

\subsection{MoBLLM Framework}
\begin{figure}[ht]
\centerline{\includegraphics[width=1\textwidth]{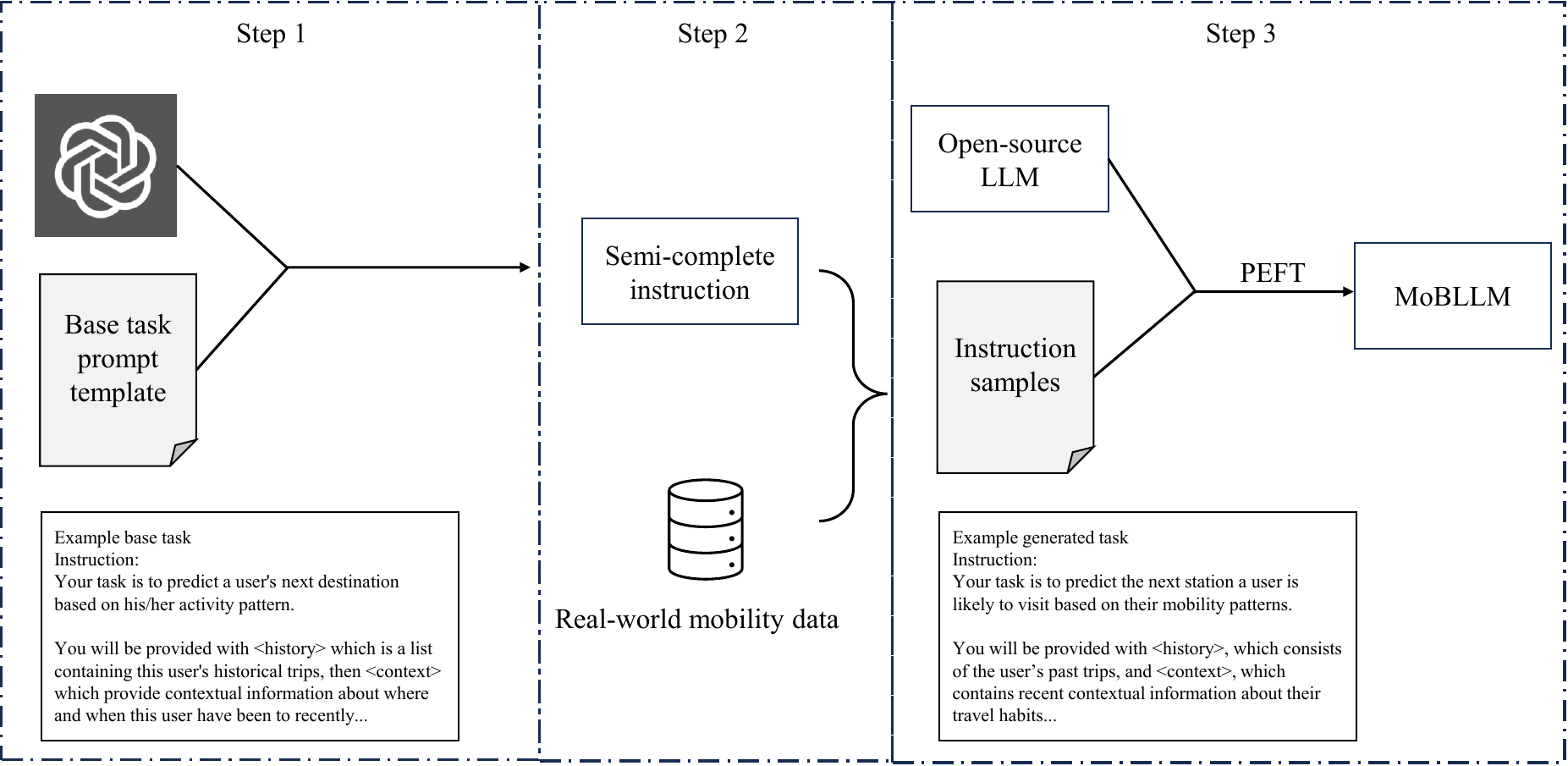}}
\caption{The overview of the proposed MoBLLM framework. Given 4 base task prompt templates, Step 1 generates multi-style semi-complete instructions without any mobility data by prompting advanced commercial LLMs such as GPT-4o mini. Step 2 constructs the instruction dataset by randomly sampling and assembling real-world mobility data with the semi-complete instructions. Step 3 obtains the MoBLLM model (for prediction) by a standard supervised fine-tuning pipeline that uses PEFT approach to train an open-source LLM on the instruction dataset.}
\label{fig_frame}
\end{figure}

Figure~\ref{fig_frame} shows the framework of foundational MoBLLM model, focusing on the fine-tuning pipeline that consists of three main steps. It is basically using a top-tier versatile commercial LLM as the teacher model to instruct a customized and dedicated student model based on open-source LLMs. Specifically, first it uses advanced commercial LLMs (e.g., OpenAI's GPT series) for fine-tuning by generating multi-style semi-complete instructions based on predefined base task prompt templates. These instructions are then combined with real-world mobility data to construct complete instruction samples. Finally, PEFT methods are applied to train an open-source LLM, such as Meta's LLaMA series~\citep{touvron2023llama}, on the instruction dataset for mobility prediction tasks. The resulting fine-tuned model is referred to as MoBLLM. It can handle multiple individual mobility prediction tasks by prompting necessary data context and guidance as shown in Figure \ref{fig_templ}.

Building upon the zero-shot prompting strategies of LLM-Mob \citep{Wang_2023}, which leveraged the general reasoning capabilities of commercial LLMs for mobility prediction, our work addresses a key limitation: these general-purpose models are not specialized for individual mobility, which constrains their predictive performance. While supervised fine-tuning methods such as PEFT provide a pathway to domain adaptation \citep{lu2025fine}, they also risk catastrophic forgetting---where the model loses its ability to generalize to contexts beyond its fine-tuning dataset \citep{sun2024amuro}. This raises concerns that fine-tuned LLMs may fail to transfer effectively to mobility data with distinct contexts (e.g., other cities or user populations) or structural types (e.g., different data formats or temporal resolutions).

To mitigate this risk and ensure robust transferability, we construct the instruction dataset with two deliberate strategies. First, we incorporate multi-type mobility data from multiple cities, enabling the LLM to capture structural patterns that are invariant across heterogeneous sources. Second, we normalize all location labels in each dataset to a universal integer index (0 to $M-1$, where $M$ is the number of unique locations). This abstraction intentionally removes dataset-specific semantics, compelling the model to focus on the underlying mechanisms of mobility dynamics---such as temporal regularities, sequential dependencies, and routine stability---rather than memorizing spatial identifiers. In doing so, MoBLLM leverages the compositional and sequential reasoning capacities of LLMs to internalize a generalized representation of human movement behavior. Although this abstraction may reduce sensitivity to spatial semantics, it substantially enhances the model’s ability to generalize across datasets with different spatial and contextual properties, which is the defining property of a foundational mobility model.

From a theoretical perspective, the generalization capability of MoBLLM can be interpreted through the lens of domain-invariant representation learning \citep{muandet2013domain, arjovsky2019invariant}. By abstracting away spatial semantics through normalized indices, MoBLLM encourages the LLM to encode mobility sequences in a latent behavioral space that is less dependent on city-specific coordinate systems or naming conventions. In this space, prediction is guided primarily by learned transition dynamics and temporal regularities rather than explicit spatial memorization. This interpretation is conceptually aligned with ideas from causal and invariant representation learning \citep{scholkopf2021toward}, in the sense that the model aims to capture behaviorally stable patterns—such as daily routines and temporal dependencies—that tend to persist across contexts. However, we emphasize that this remains a conceptual lens: no formal causal or invariance guarantees are claimed. Consequently, adaptation to unseen datasets is viewed as distributional alignment within this latent behavioral space, rather than proof of causal mechanism discovery.

\subsection{Prompt Template for Individual Mobility Prediction}
To prompt LLMs to predict individual mobility in a batch processing manner, we employ an efficient template based on zero-shot CoT, following similar structures commonly adopted in recent literature \citep{Wang_2023, qin2025lingotrip, liang2024exploring}. The zero-shot CoT enhances LLMs to capture the spatiotemporal mobility patterns from historical data. As shown in Figure \ref{fig_templ}, the prompt template includes five components: task definition, data description, thinking guidance, output format, and data inputs. 

First, the template briefly introduces the prediction tasks according to the definitions of base tasks in Section \ref{sec_task}. To improve LLMs' capability of understanding the context of mobility data, we provide detailed descriptions about the formatted data, which are adjusted due to the practical data usage of the corresponding tasks. The output format regulates the LLM’s output by asking it to provide its prediction and reason, facilitating quick processing of the text output. Such an approach of asking explanation not only improves LLMs' prediction performance in mobility prediction but also increases the interpretability and reliability of the model \citep{Wang_2023}. The thinking guidance is the key to carrying out the LLMs' zero-shot learning in individual mobility prediction. It guides LLMs to capture mobility patterns from different time scales, i.e., the first and second aspects of thinking guidance shown in Figure \ref{fig_templ}. Then the third aspect reminds LLMs to learn the semantic information from the trips' spatiotemporal context.

\subsection{Instruction Data Generation}
\label{inst_data_gen}
\begin{figure}[ht]
\centerline{\includegraphics[width=1\textwidth]{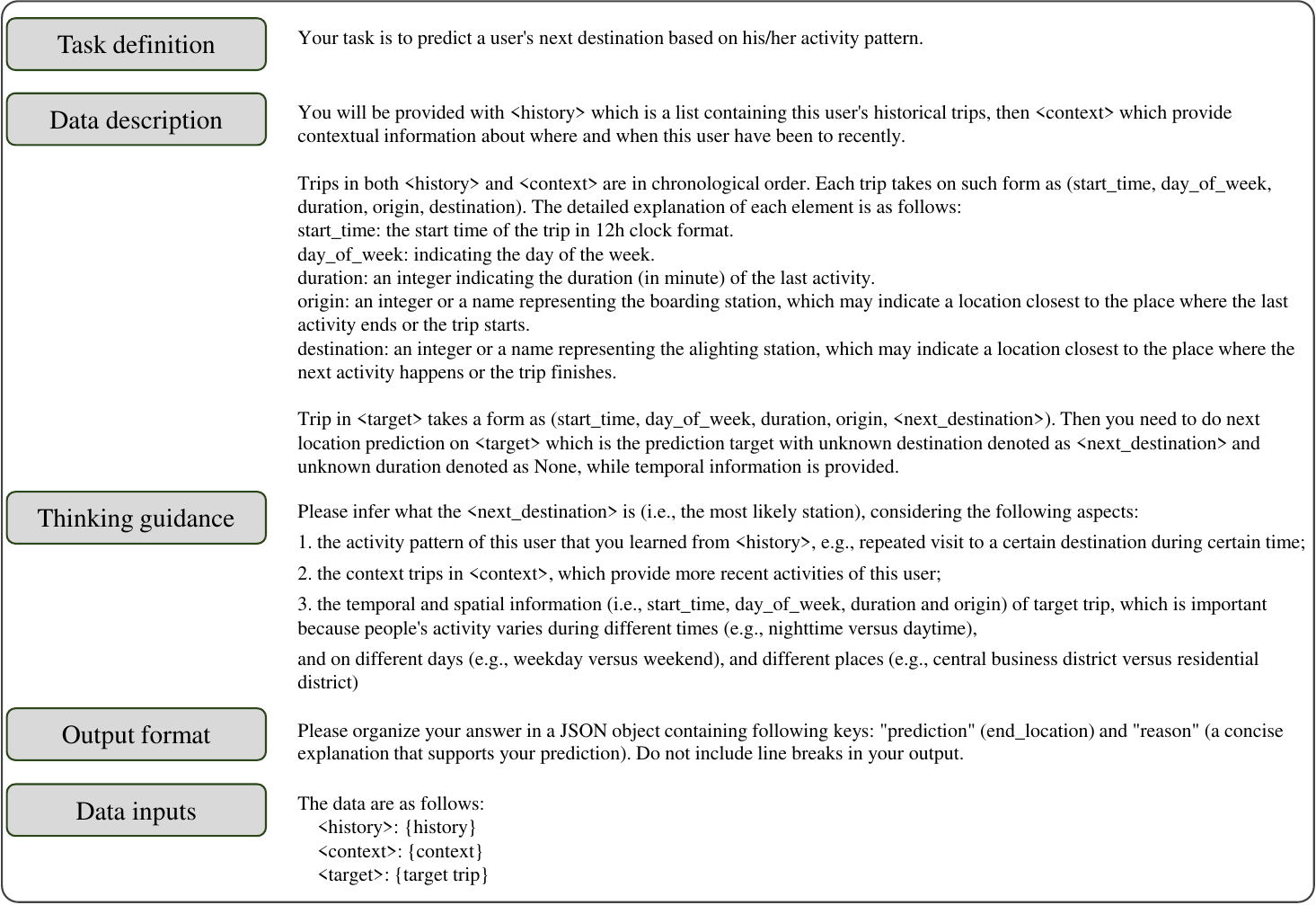}}
\caption{The example of instruction data. For the data input, $<$history$>$ and $<$context$>$ are the user's mobility sequences respectively, containing long-term and recent stays or trips. $<$target$>$ includes the information about current states except for location, e.g., the tap-out station and time of the last trip.}
\label{fig_templ}
\end{figure}

Since most current LLMs are initially designed for dealing with various natural language tasks, there is a non-trivial mismatch between the original training objective of LLMs and our studied individual mobility prediction problem. To address the mismatch, we adopt the instruction based fine-tuning approach, which is commonly applied to enhance the capabilities and controllability of LLMs in other domain-specific tasks \citep{zhang2023instruction}. The key to the approach is to construct an instruction dataset for LLMs' fine-tuning. Each sample in the dataset can be concisely represented as an $(Instruction, Output)$ pair. $Instruction$ is a kind of prompt for instructing LLMs to specialize in the studied problem during the fine-tuning process, as shown in Figure \ref{fig_templ}. The $Output$ denotes the desired output that follows the $Instruction$.

\begin{figure}[]
\centerline{\includegraphics[width=1\textwidth]{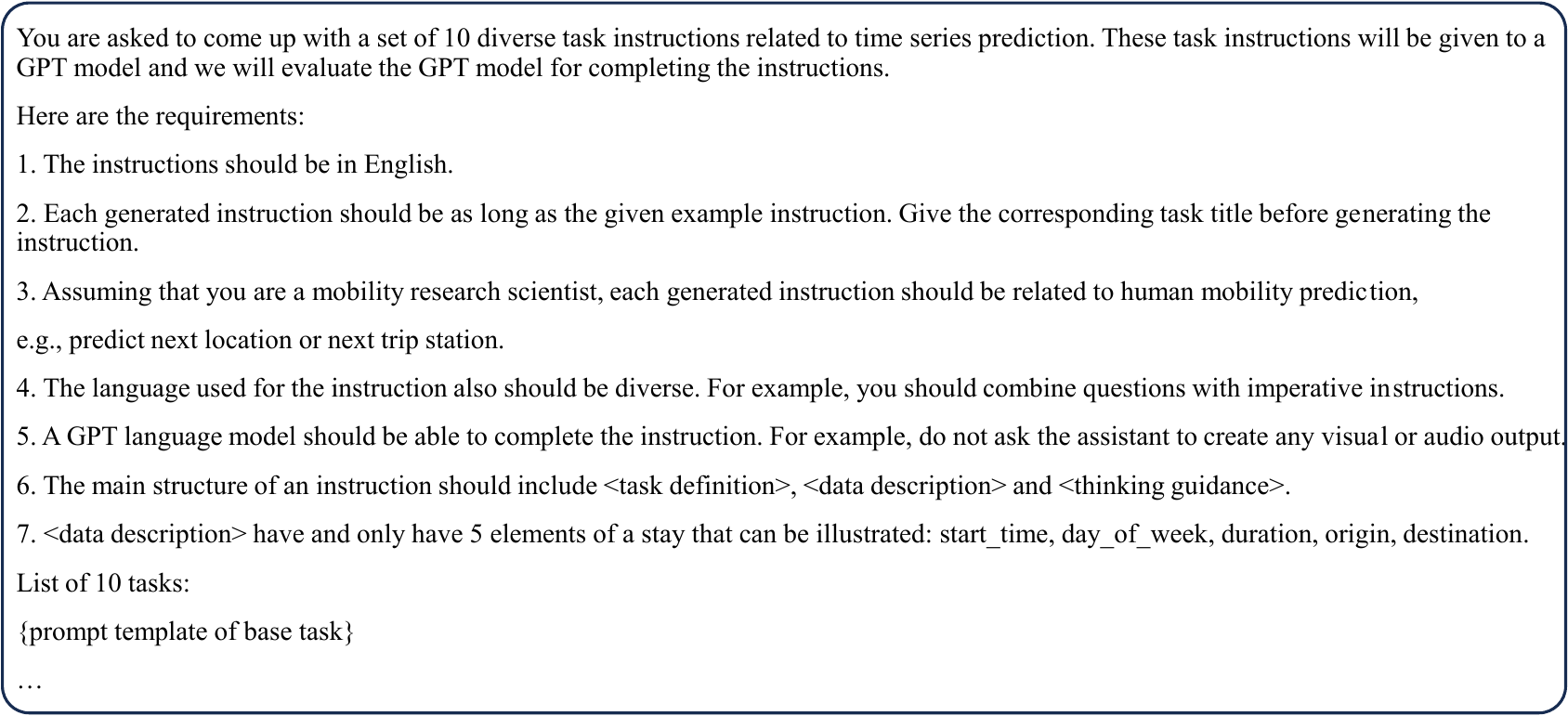}}
\caption{The prompt example to generate multi-style instructions. A prompt template of the base task is put at the end, prompting LLMs to learn the style.}
\label{fig_genPrompt}
\end{figure}

In practical application scenarios, users have different styles in writing prompts. The LLMs fine-tuned by single-style instructions have difficulty adapting to varying prompts, which probably leads to model performance reduction. To address this issue, we adopt an Alpaca-style prompt \citep{taori2023alpaca} to distill different expressions describing individual mobility prediction tasks to generate sufficient multi-style task instructions as the examples shown in Figure \ref{fig_frame}. Typically, distillation imparts knowledge and cognitive abilities from a competent teacher model to a simpler and more computationally efficient student model \citep{zhang2023instruction}. Figure \ref{fig_genPrompt} shows the CoT-based prompt for generating multi-style instructions with three main parts (see \ref{apx_instru_eg} for generated examples). First, the LLMs should know how to generate diverse instructions related to time series prediction. Then, the requirements assist LLMs to work like professional mobility researchers to focus on the goal of human mobility prediction. In particular, it is necessary to limit the generation scope to text and obligatorily consider the generated instruction structure in the last three requirements. Such structural information can guide LLMs to generate CoT-based task instructions, which have been demonstrated to improve LLMs' mobility prediction performance by previous studies \citep{Wang_2023, liang2024exploring, feng2024move, qin2025lingotrip}. To ensure the data authenticity, only the task definition, data description, and thinking guidance in Figure \ref{fig_templ} are generated with diverse styles to get semi-complete instructions in Figure \ref{fig_frame}. Then it combines with a piece of real-world mobility data from the database to get an $Instruction$. To align with the required prediction results of the task prompt, the corresponding $Output$ given an $Instruction$ is set as $\{prediction: place\_id\}$ where the $place\_id$ is the ground truth of the next location or origin / destination station defined in the Problem Definition.

\subsection{Parameter Efficient Fine-Tuning}

Fine-tuning is a widely adopted approach for adapting LLMs to domain-specific datasets. It leverages the pre-trained knowledge embedded in LLMs to enhance task-specific performance while requiring relatively limited data and computational resources \citep{2024ultimate}. However, conventional approaches such as full-parameter fine-tuning demand substantial Graphics Processing Unit (GPU) memory and computational overhead, making them impractical for use on resource-constrained devices or small-scale workstations. In contrast, parameter-efficient fine-tuning (PEFT) methods significantly reduce memory requirements while maintaining high adaptation performance \citep{han2024parameter}. One of the most prominent PEFT techniques is Low-Rank Adaptation (LoRA) \citep{hu2022lora}, which has been extensively used to fine-tune LLMs in mobility-related applications \citep{li2024limp, gong2024mobility}.

\begin{figure}[ht]
\centerline{\includegraphics[width=0.3\textwidth]{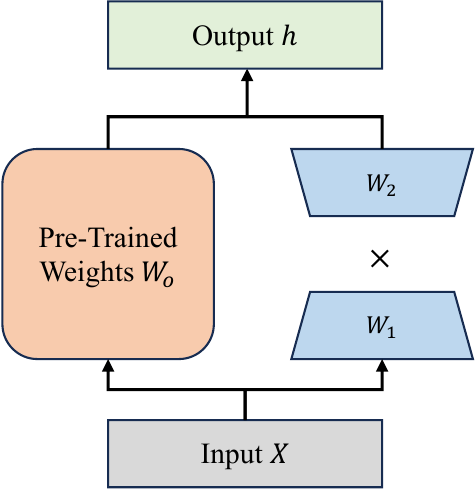}}
\caption{The framework of LoRA.}
\label{fig_lora}
\end{figure}

Figure~\ref{fig_lora} shows that LoRA introduces two trainable matrices, $W_1 \in \mathbb{R}^{d \times r}$ and $W_2 \in \mathbb{R}^{r \times k}$, where the rank $r$ is significantly smaller than $\min(d, k)$. Given an input $X \in \mathbb{R}^{d \times k}$, the original pre-trained weight matrix $W_0$ remains fixed throughout the fine-tuning process. This matrix encodes general-purpose knowledge acquired during large-scale pretraining. In contrast, $W_1$ and $W_2$ are lightweight adapters that capture task-specific patterns—in this case, individual mobility behavior. The adapted representation used for downstream mobility prediction tasks is computed as:
\begin{equation}
    h = W_0X + \frac{\alpha}{r} W_1W_2X,
    \label{eq_lora}
\end{equation}
where $h$ denotes the output embedding aligned with mobility-related semantics in $X$. Here, $\alpha$ is a tunable scaling factor used to control the contribution of the low-rank update. Instead of updating the full weight matrix, LoRA introduces a low-rank correction term $W_1W_2$ on top of $W_0$, allowing the model to efficiently adapt without retraining all parameters. Since the number of parameters in $W_1$ and $W_2$ is orders of magnitude smaller than in $W_0$, this approach significantly reduces GPU memory requirements. Empirical studies report that LoRA can achieve up to a 3 times reduction in memory usage while maintaining competitive performance \citep{hu2022lora}, making it well-suited for fine-tuning LLMs on mobility datasets under limited resources.

In the fine-tuning phase, the individual mobility prediction task is framed as a next-token generation task. With the instruction dataset constructed by Section \ref{inst_data_gen}, the PEFT method can tune an open-source LLM's parameters by optimizing the following objective function in a supervised way:
\begin{equation}
    L(\theta) = -\mathbb{E}_{(x,y)\in \mathbb{D}}[\sum_{i=1}^n log_{\theta}\space P(y_i|x;y_{1:i-1})],
\end{equation}
where $\mathbb{D}$ denotes the instruction dataset and $\theta$ represents the trainable parameters in PEFT method (i.e., $W_1$ and $W_2$ in Equation \ref{eq_lora}). Each $(x,y)$ represents a pair of $(Instruction, Output)$ in the instruction dataset. The $y_{1:i-1}$ and $y_i$ respectively denote the previously generated tokens and the ground truth of the next token.

\section{Case Studies}
\label{sec4}
\subsection{Data Description and Performance metrics}
We aim to systematically validate the MoBLLM model capability in generalizing to diverse prediction tasks across varying time scales, spatial scales and conditional scenarios. To do that, we conduct case studies using six mobility datasets, i.e., Geolife \footnote{https://www.microsoft.com/en-us/research/publication/geolife-gps-trajectory-dataset-user-guide/}, FSQ-NYC, FSQ-TKY \citep{yang2014modeling}, FSQ-Global\footnote{https://sites.google.com/site/yangdingqi/home/foursquare-dataset} \citep{yang2016participatory}, Hong Kong (HK) AFC data, Hangzhou (HZ) AFC data \footnote{https://tianchi.aliyun.com/dataset/21904}. The detailed statistics of these datasets are presented in Table \ref{tab_data}. According to Tasks 3 and 4 for predicting origin and destination, each AFC data can be processed to obtain two subsets, such as HK-ORI and HK-DEST. 

The MoBLLM model is obtained by following the pipeline in Figure \ref{fig_frame} that fine-tunes a pre-trained LLaMA-3.1-8B-Instruct \footnote{https://huggingface.co/meta-llama/Llama-3.1-8B-Instruct} on the instruction dataset. We set MoBLLM's temperature parameter to zero during inference to guarantee deterministic output across all case studies. In our transferability experiments, the training data were sampled from four datasets (Geolife, FSQ-NYC, HK-ORI and HK-DEST), while testing was conducted on four different datasets (FSQ-TYK, FSQ-Global, HZ-ORI, and HZ-DEST). As shown in Table \ref{tab_data}, these datasets differ significantly in terms of data type (GPS trajectory, check-in, AFC), geographical context (Beijing, New York, Tokyo, Hong Kong, Hangzhou, and 415 global cities), and spatial coverage (ranging from dozens to over 100,000 unique locations). These differences suggest that the transfer experiments indeed involve substantial cross-dataset heterogeneity rather than trivial similarities.

\begin{table}[]
\centering
\scriptsize
\caption{The descriptions of datasets.}
\begin{tabular}{llllllll}
\hline
\textbf{Dataset} & \textbf{Type}  & \textbf{Users} & \textbf{Test Samples} & \textbf{District}          & \textbf{Locations} & \textbf{Time Span} & \textbf{Task Type} \\ \hline
Geolife          & GPS trajectory & 45             & 3459                  & Beijing                    & 1145               & 2007/04-2012/08    & task 1             \\
FSQ-NYC          & Check-in       & 535            & 12141                 & New York                   & 3752               & 2011/10-2012/02    & task 2             \\
FSQ-TYK          & Check-in       & 450            & 11250                 & Tokyo                      & 15785              & 2012/04-2013/02    & task 2             \\
FSQ-Global       & Check-in       & 1500           & 43500                 & \begin{tabular}[c]{@{}l@{}}415 cities in 77\\ countries.\end{tabular} & 112032             & 2012/04-2013/09    & task 2             \\
HK-ORI           & AFC            & 500            & 5260                  & Hong Kong                  & 92                 & 2018/01-2018/03    & task 3             \\
HK-DEST          & AFC            & 500            & 5260                  & Hong Kong                  & 92                 & 2018/01-2018/03    & task 4             \\
HZ-ORI           & AFC            & 1101           & 15296                 & Hangzhou                   & 81                 & 2019/01-2019/02    & task 3             \\
HZ-DEST          & AFC            & 1101           & 15296                 & Hangzhou                   & 81                 & 2019/01-2019/02    & task 4             \\ \hline
\end{tabular}
\label{tab_data}
\end{table}

In the case studies, we employ the following two metrics to evaluate the predictive performance of models:

\begin{itemize}
\item \textbf{Accuracy}: measures if the next location's prediction is correct compared with the ground truth of the next visited location. We use $ACC (\%)$ in the following parts to denote this metric that measures the ratio of correct predictions in the test set, which is formulated as follows:
\begin{equation}
    ACC = \frac{T}{T + N} \times 100\%,
\end{equation}
where $T$ and $N$ are the number of correctly and incorrectly predicted samples, respectively.

\item \textbf{Weighted F1} measures the model's performance in predicting frequently visited locations to avoid the issue of imbalance of individual's visited locations.  which is denoted as the abbreviation $F1$ in the following parts. 
\begin{equation}
    F1 = \sum_{i\in D} \frac{s_i}{S} \cdot \frac{2TP_i}{2TP_i + FP_i + FN_i},
\end{equation}
where $TP_i$, $FP_i$ and $FN_i$ are respectively the number of true positive, false positive and false negative samples given a ground truth location $l_i$. The $s_i$ represents samples of ground truth $l_i$ and the $S$ denotes the size of the dataset $D$. We use the F1 score weighted by the number of visits to emphasize the overall performance of the model in the most important locations.
\end{itemize}

\subsection{PEFT methods and parameters}
\label{sec_peft}
To adapt open-source LLMs to real-world human mobility prediction tasks under resource constraints, we apply LoRA and several advanced variants in fine-tuning and compare their efficiency. These include OLoRA \citep{buyukakyuz2024olora}, EVA \citep{paischer2024one}, PiSSA \citep{meng2024pissa}, LoftQ \citep{li2023loftq}, LoRA+ \citep{hayou2024lora+}, rsLoRA \citep{kalajdzievski2023rank}, and QLoRA \citep{dettmers2023qlora}. 

There are two important hyperparameters in Equation \eqref{eq_lora} that have a significant impact on fine-tuning. They are the rank $r$ and scaling factor $\alpha$, which control the updated parameter size and the ratio of specific knowledge influencing the fine-tuning process. Their specific values used in the PEFT methods in Table \ref{tab_peft} are referred to in the corresponding literature, considering the trade-off between GPU memory, fine-tuning time, and model performance. All PEFT methods used in this research are based on a pre-trained LLaMA-3.1-8B-Instruct except the LoftQ starting with a 4-bit and 64-rank checkpoint of LLaMA-3-8B-Instruct \footnote{https://huggingface.co/LoftQ/Meta-Llama-3-8B-Instruct-4bit-64rank} provided. About 100K training samples are used for fine-tuning and the batch size is set to 3 for all the PEFT methods following common practices in LoRA-based fine-tuning studies that adopt very small batch sizes due to GPU memory constraints \citep{li2024limp}. All the model training and experimental evaluations are conducted on a workstation with Intel(R) Xeon(R) w7-3465X CPU, 4 NVIDIA RTX 6000 Ada GPUs, and 512-Gb memory.





As shown in Table \ref{tab_peft}, the OLoRA achieves the best performance on average with the highest GPU memory consumption but the lowest fine-tuning time. The QLoRA has the lowest performance, though it needs the least GPU memory for fine-tuning because it represents weights and activations with lower-precision data types by a 4-bit quantization technique. For the following experiments, we adopt the checkpoint trained by OLoRA to compare the MoBLLM with other baseline models.

\begin{table}[]
\small
\centering
\caption{The performance of MoBLLM model with different PEFT methods. The MEM denotes the GPU memory (Gb) and Time denotes the fine-tuning time (hour).}
\begin{tabular}{lllllllllll}
\hline
\multirow{2}{*}{\textbf{PEFT}} & \multicolumn{2}{l}{\textbf{Geolife}}                          & \multicolumn{2}{l}{\textbf{FSQ-NYC}}                          & \multicolumn{2}{l}{\textbf{HK-ORI}}                           & \multicolumn{2}{l}{\textbf{HK-DEST}}                          & \multirow{2}{*}{\textbf{MEM}} & \multirow{2}{*}{\textbf{Time}} \\ \cline{2-9}
                      & \multicolumn{1}{l}{\textbf{ACC}}            & \textbf{F1}              & \multicolumn{1}{l}{\textbf{ACC}}            & \textbf{F1}              & \multicolumn{1}{l}{\textbf{ACC}}            & \textbf{F1}              & \multicolumn{1}{l}{\textbf{ACC}}            & \textbf{F1}              &                      &                       \\ \hline
LoRA                  & \multicolumn{1}{l}{58.31}          & 0.5385          & \multicolumn{1}{l}{41.58}          & 0.3687          & \multicolumn{1}{l}{84.52}          & 0.8447          & \multicolumn{1}{l}{70.3}           & 0.6972          & 50                   & 30                    \\
OLoRA                 & \multicolumn{1}{l}{58.51}          & 0.5394          & \multicolumn{1}{l}{41.21}          & 0.3651          & \multicolumn{1}{l}{\textbf{87.86}} & \textbf{0.8784} & \multicolumn{1}{l}{\textbf{70.93}} & \textbf{0.7044} & 60.4                 & 23                    \\
EVA                   & \multicolumn{1}{l}{57.13}          & 0.5251          & \multicolumn{1}{l}{40.51}          & 0.3568          & \multicolumn{1}{l}{87.43}          & 0.8743          & \multicolumn{1}{l}{70.23}          & 0.6979          & 54.3                 & 25.5                  \\
PiSSA                & \multicolumn{1}{l}{53.89}          & 0.4829          & \multicolumn{1}{l}{35.21}          & 0.3053          & \multicolumn{1}{l}{62.39}          & 0.6259          & \multicolumn{1}{l}{41.77}          & 0.3972          & 60                   & 32                    \\
LoftQ                 & \multicolumn{1}{l}{58.5}           & 0.537           & \multicolumn{1}{l}{40.92}          & 0.3611          & \multicolumn{1}{l}{87.46}          & 0.8745          & \multicolumn{1}{l}{70.48}          & 0.6995          & 42.1                 & 15.2                 \\
LoRA+                 & \multicolumn{1}{l}{58.14}          & 0.5346          & \multicolumn{1}{l}{39.86}          & 0.3487          & \multicolumn{1}{l}{63.75}          & 0.6348          & \multicolumn{1}{l}{41.52}          & 0.3861          & 54.7                 & 32                    \\
rsLoRA                 & \multicolumn{1}{l}{\textbf{59.01}} & \textbf{0.5414} & \multicolumn{1}{l}{\textbf{41.55}} & \textbf{0.3678} & \multicolumn{1}{l}{87.48}          & 0.8749          & \multicolumn{1}{l}{68.56}          & 0.6805          & 58.5                 & 32                    \\
QLoRA                  & \multicolumn{1}{l}{38.48}          & 0.3348          & \multicolumn{1}{l}{18.15}          & 0.1355          & \multicolumn{1}{l}{40.08}          & 0.3883          & \multicolumn{1}{l}{42.7}           & 0.4028          & 46.4                 & 37.4                  \\ \hline
\end{tabular}

\label{tab_peft}
\end{table}

\subsection{Baselines}
\label{sec_basel}
To benchmark the performance of MoBLLM, we compare it with 4 baseline models as follows:

\begin{itemize}
\item \textbf{DeepMove} \citep{Feng2018} uses recurrent networks with an attention mechanism to capture multi-scale temporal dependencies of mobility patterns.

\item \textbf{MobTCast} \citep{xue2021mobtcast} is based on a transformer encoder network to leverage temporal, semantic, social, and geographical contexts from historical mobility sequences to predict the next location.

\item \textbf{MHSA} \citep{hong2023context} uses a multi-head self-attentional network to infer an individual's next location by considering historical location visits, visit time, activity duration, as well as their surrounding land use information.

\item \textbf{LLM-Mob} \citep{Wang_2023} is an ICL-based model to first use LLMs to predict future human mobility by zero-shot CoT.

\end{itemize}

All the deep learning models DeepMove, MobTCast and MHSA are respectively trained and tested on each dataset with the same sample size as the MoBLLM. The MoBLLM model is fine-tuned on a mixed dataset combined with Geolife, FSQ-NYC, HK-ORI, and HK-DEST where each sample is assembled with an individual historical mobility sequence and instruction context generated by GPT-4o mini of Azure OpenAI with temperature as 0.3. It consumes about 1.2M tokens of generated instructions at a friendly cost of 0.7 dollars. The LLM-Mob model also adopts GPT-4o mini with temperature as zero to test its performance on the datasets.

\subsection{Prediction Performance Comparison}
Table \ref{tab_comp} shows the results of comparing the MoBLLM moel with baselines. The MoBLLM achieves the state-of-the-art performance over all the baseline models on all datasets. The deep learning models DeepMove, MobTcast, and MHSA can perform better only on HK-ORI and HK-DEST than the LLM-Mob model. Compared with Geolife and FSQ-NYC, which contain various activity patterns with high sparsity, these AFC datasets have dense mobility patterns like periodical commuting. These deep learning models can easily learn dense patterns in training but have difficulty capturing sparse patterns, while the MoBLLM can adapt to different types of datasets regardless of sparse or dense patterns. The results of comparing the two LLM-based methods demonstrate that open-source LLMs can obtain substantial performance improvement by PEFT fine-tuning (better than commercial ones). 

\begin{table}[ht]
\small
\centering
\caption{The performance comparison between the MoBLLM and baseline models}
\begin{tabular}{lllllllll}
\hline
\multirow{2}{*}{\textbf{Model}} & \multicolumn{2}{l}{\textbf{Geolife}}        & \multicolumn{2}{l}{\textbf{FSQ-NYC}}        & \multicolumn{2}{l}{\textbf{HK-ORI}}         & \multicolumn{2}{l}{\textbf{HK-DEST}}         \\ \cline{2-9} 
                       & \multicolumn{1}{l}{\textbf{ACC}}   & \textbf{F1}     & \multicolumn{1}{l}{\textbf{ACC}}   & \textbf{F1}     & \multicolumn{1}{l}{\textbf{ACC}}   & \textbf{F1}     & \multicolumn{1}{l}{\textbf{ACC}}   & \textbf{F1}     \\ \hline
DeepMove               & \multicolumn{1}{l}{26.10} & 0.1890 & \multicolumn{1}{l}{19.30} & 0.1550 & \multicolumn{1}{l}{82.81} & 0.827  & \multicolumn{1}{l}{65.45} & 0.6450 \\
MobTCast               & \multicolumn{1}{l}{29.50} & 0.1730 & \multicolumn{1}{l}{20.20} & 0.1660 & \multicolumn{1}{l}{83.28} & 0.8324 & \multicolumn{1}{l}{65.49} & 0.6478 \\
MHSA                   & \multicolumn{1}{l}{31.40} & 0.2180 & \multicolumn{1}{l}{20.20} & 0.1490 & \multicolumn{1}{l}{83.55} & 0.8349 & \multicolumn{1}{l}{66.00} & 0.6554 \\
LLM-Mob                & \multicolumn{1}{l}{45.10} & 0.4040 & \multicolumn{1}{l}{27.40} & 0.2210 & \multicolumn{1}{l}{62.81} & 0.6494 & \multicolumn{1}{l}{54.23} & 0.5334 \\
MoBLLM                  & \multicolumn{1}{l}{\textbf{58.51}} & \textbf{0.5394} & \multicolumn{1}{l}{\textbf{41.21}} & \textbf{0.3651} & \multicolumn{1}{l}{\textbf{87.86}} & \textbf{0.8784} & \multicolumn{1}{l}{\textbf{70.93}} & \textbf{0.7044} \\ \hline
\end{tabular}

\label{tab_comp}
\end{table}

\subsection{Transferability Performance}
\emph{Transferability} measures the model's performance in adapting to data sources that are significantly different from the training data. Based on the checkpoint trained by OLoRA in Section \ref{sec_peft}, the MoBLLM is tested its transferability on the datasets FSQ-TYK, FSQ-Global, HZ-ORI and HZ-DEST as described in Table \ref{tab_data}, which are unseen in the fine-tuning. Due to the inconsistency of predictable locations between training data and the transferability test data, the conventional deep learning models DeepMove, MobTCast, and MHSA are unable to be applied to other cities' mobility data. Therefore, we only use the LLM-Mob deployed with GPT-4o mini as the baseline in this experiment. To avoid the disturbance caused by different prompts, the MoBLLM adopts the same prompt as LLM-Mob, such as the example in Figure \ref{fig_templ}.

The results are presented in Table \ref{tab_transfer}, where the open-source MoBLLM model significantly outperforms the general-purpose commercial LLM. Both models perform worse on the check-in datasets than the AFC datasets since the former exhibit substantially higher sparsity in the set of possibly visited locations as shown in Table \ref{tab_data}. This issue is particularly pronounced in FSQ-Global, which contains over 112,000 unique locations across 415 cities, resulting in extreme heterogeneity and long-tailed distributions. Such characteristics make prediction more difficult and contribute to the relatively low accuracy observed. Nevertheless, MoBLLM still demonstrates much stronger capabilities than LLM-Mob in capturing sparse mobility patterns, indicating that the main limitation arises from data sparsity and distribution complexity rather than the MoBLLM architecture itself. It demonstrates that an open-source LLM, i.e., LLaMA-3.1-8B-Instruct used in the case studies, can become an expert in different mobility prediction tasks by the proposed fine-tuning framework. Moreover, the MoBLLM shows a significant superiority in cost efficiency, mainly produced by generating instructions for training, which is about 60 times cheaper than a commercial LLM.

\begin{table}[]
\centering
\caption{The transferability and cost comparison between GPT-4o mini and MoBLLM on four datasets. The cost denotes the total expense incurred in the test.}
\begin{tabular}{llllllllll}
\hline
\multirow{2}{*}{\textbf{Model}} & \multicolumn{2}{l}{\textbf{HZ-ORI}}                           & \multicolumn{2}{l}{\textbf{HZ-DEST}}                          & \multicolumn{2}{l}{\textbf{FSQ-TYK}}                         & \multicolumn{2}{l}{\textbf{FSQ-Global}}                       & \multirow{2}{*}{\textbf{Cost(\$)}} \\ \cline{2-9}
                       & \multicolumn{1}{l}{\textbf{ACC}}            & \textbf{F1}              & \multicolumn{1}{l}{\textbf{ACC}}            & \textbf{F1}              & \multicolumn{1}{l}{\textbf{ACC}}           & \textbf{F1}              & \multicolumn{1}{l}{\textbf{ACC}}            & \textbf{F1}              &                           \\ \hline
LLM-Mob            & \multicolumn{1}{l}{74.03}          & 0.7473          & \multicolumn{1}{l}{69.26}          & 0.6901          & \multicolumn{1}{l}{16.2}          & 0.1171          & \multicolumn{1}{l}{16.26}          & 0.1066          & 43.8                      \\
MoBLLM                 & \multicolumn{1}{l}{\textbf{89.24}} & \textbf{0.8921} & \multicolumn{1}{l}{\textbf{80.15}} & \textbf{0.7998} & \multicolumn{1}{l}{\textbf{28.7}} & \textbf{0.2528} & \multicolumn{1}{l}{\textbf{25.42}} & \textbf{0.2085} & 0.7                       \\ \hline
\end{tabular}

\label{tab_transfer}
\end{table}

\subsection{Robustness under Different Prediction Situations}
To investigate the robustness under different situations (events, incidents), we further estimate the model prediction performance in the Hong Kong AFC data (mainly due to the wide data availability for the test). As shown in Table \ref{tab_adap_data}, the adaptation datasets involve individual trips occurring in metro network change (MNC), policy intervention (PI), special events (SE), and metro incidents (MI). Besides, each of the four datasets can be divided into 2 subsets according to origin prediction (ORI) and destination prediction (DEST). All robustness validation results are shown in Table \ref{tab_ex_scenarios}, where the results for HK-ORI and HK-DEST under regular scenarios are also listed for reference. The $\Delta$ACC denotes the performance difference between regular and irregular scenario, providing a direct measure of robustness across models.

Overall, the MoBLLM model has robust performance in different prediction situations and significantly outperforms all the baseline models. The deep learning based models have huge performance reduction in PI-ORI and PI-DEST. Especially in PI-DEST, their performance reduces by over 10\% compared with the performance in HK-DEST, while the LLM-based models only have a margin reduction of about 3\%. The probable reason is that the PI datasets are much different from the HK-ORI and HK-DEST used for training the deep learning models. Deep learning models have low transferability, which leads to significant performance variation in situations where the input data is inconsistent with the training data to some extent.

We further analyze the robustness of the MoBLLM model in each prediction situation. In datasets MNC and PI that have fewer stations than the training data, the MoBLLM model performance has an average decrease by about 4\% on ACC for origin prediction compared to the performance in HK-ORI. The MoBLLM model performs worse in PI than MNC for both origin and destination prediction because there exist policy interventions of price discounts that probably cause a part of passengers to change their travel behaviors. In the datasets SE and MI that have special events and incidents influencing the demands, the MoBLLM model has a margin performance reduction of about 3\% for the origin prediction. It is because a small number of passengers may change their travel choices. For destination prediction on these two datasets, the MoBLLM even has better performance than the one on HK-DEST, which implies that the special events may not significantly influence the destination of existing trips. Therefore, the MoBLLM model has a good robustness to adapt to different prediction situations covering metro network change, policy disturbances, special events, and incidents.

\begin{table}[]
\small
\centering
\caption{The descriptions of adaptation datasets.}
\begin{tabular}{lllll}
\hline
\textbf{Dataset} & \textbf{User Number} & \textbf{Test Samples} & \textbf{Time Span}       & \textbf{Description}                                                                                                                                                                                      \\ \hline
MNC     & 898         & 6318         & 2016/08-2016/11 & \begin{tabular}[c]{@{}l@{}}Compared with training data,\\ it has less stations.\end{tabular}                                                                                                                                             \\ \hline
PI      & 844         & 5791         & 2014/07-2014/08 & \begin{tabular}[c]{@{}l@{}}Compared with training data,\\ it has less stations and fare \\ discount in morning peak.\end{tabular}                                                                  \\ \hline
SE      & 1657        & 5010         & 2018/01-2018/03 & \begin{tabular}[c]{@{}l@{}}Compared with training data,\\ it has the same number of stations \\ but all the samples are selected \\ from 35 days with live concerts.\end{tabular}                     \\\hline
MI      & 2919        & 4975         & 2018/01-2018/02 & \begin{tabular}[c]{@{}l@{}}Compared with training data,\\ it has the same number of stations \\
but all the samples are selected\\ from 7 incident days with at least\\ 10-minute delay.\end{tabular} \\ \hline
\end{tabular}
\label{tab_adap_data}
\end{table}

\begin{table}[]
\small
\centering
\caption{Model performance comparison under different scenarios. Row '$\Delta$ACC' represents the improvement percentage relative to the 'ACC' rows of HK-ORI or HK-DEST according to the corresponding tasks.}
\begin{tabular}{cllllll}
\hline
\multicolumn{2}{c}{\textbf{Dataset}}           & \textbf{DeepMove} & \textbf{MobTCast} & \textbf{MHSA}   & \textbf{LLM-Mob} & \textbf{MoBLLM} \\ \hline 
\multirow{2}{*}{HK-ORI}   & ACC       & 82.81    & 83.28    & 83.55  & 62.81   & 87.86  \\
                          & $\Delta$ACC & 0.00     & 0.00     & 0.00   & 0.00    & 0.00   \\
\multirow{2}{*}{MNC-ORI}  & ACC       & 74.39    & 78.80    & 78.69  & 63.49   & 83.10  \\
                          & $\Delta$ACC & -10.17   & -5.38    & -5.82  & 1.08    & -5.42  \\
\multirow{2}{*}{PI-ORI}   & ACC       & 71.52    & 75.91    & 76.51  & 58.97   & 79.40  \\
                          & $\Delta$ACC & -13.63   & -8.85    & -8.43  & -6.11   & -9.63  \\
\multirow{2}{*}{SE-ORI}   & ACC       & 73.99    & 78.68    & 79.34  & 62.06   & 84.91  \\
                          & $\Delta$ACC & -10.65   & -5.52    & -5.04  & -1.19   & -3.36  \\
\multirow{2}{*}{MI-ORI}   & ACC       & 75.47    & 80.78    & 81.56  & 65.33   & 85.59  \\
                          & $\Delta$ACC & -8.86    & -3.00    & -2.38  & 4.01    & -2.58   \\ \hline
\multirow{2}{*}{HK-DEST}  & ACC       & 65.45    & 65.49    & 66.00  & 54.23   & 70.93  \\
                          & $\Delta$ACC & 0.00     & 0.00     & 0.00   & 0.00    & 0.00   \\
\multirow{2}{*}{MNC-DEST} & ACC       & 55.22    & 59.43    & 58.76  & 55.67   & 72.29  \\
                          & $\Delta$ACC & -15.63   & -9.25    & -10.97 & 2.66    & 1.92   \\
\multirow{2}{*}{PI-DEST}  & ACC       & 51.54    & 55.08    & 55.15  & 53.76   & 68.71  \\
                          & $\Delta$ACC & -21.25   & -15.90   & -16.44 & -0.87   & -3.13  \\
\multirow{2}{*}{SE-DEST}  & ACC       & 53.17    & 56.68    & 58.86  & 52.93   & 72.69  \\
                          & $\Delta$ACC & -18.76   & -13.45   & -10.82 & -2.40   & 2.48   \\
\multirow{2}{*}{MI-DEST}  & ACC       & 60.70    & 63.41    & 64.78  & 61.95   & 79.92  \\
                          & $\Delta$ACC & -7.26    & -3.18    & -1.85  & 14.24   & 12.67  \\ \hline
\end{tabular}
\label{tab_ex_scenarios}
\end{table}

\section{Conclusions}
This paper proposes the MoBLLM model for individual mobility prediction problem. It is obtained by adopting the PEFT method to tune an open-source LLM on the instruction dataset, which can be generated based on the real-world mobility data. Such fine-tuning enables the MoBLLM model to adapt to diverse individual prediction tasks across time scales, spatial scales, and conditional scenarios.

In the cases studies, six real-world mobility datasets from multiple cities are used to validate the performance of MoBLLM model. We systematically evaluate LoRA and its several advanced variants in fine-tuning to obtain the optimal MoBLLM model. Compared with the state-of-art baseline models on accuracy, the MoBLLM model averagely outperforms the best DL model (MHSA) by 50.44\% and the LLM-Mob model based on commercial LLMs by 37.30\%. Besides, the MoBLLM model also shows significant superiority in transferability and finance compared to the method based on commercial LLMs. In the robustness evaluation on 4 trip datasets representing different conditional scenarios, the MoBLLM shows the robust performance and outperforms the best benchmark (MHSA) by 14.67\% on average. These results of case studies comprehensively demonstrate the capacity of the MoBLLM model as the foundational individual mobility prediction model.

This work has explored the application of Zero-Shot CoT prompting on the fine-tuned LLMs for individual mobility prediction. It is important to note, however, that the sequential nature of CoT reasoning introduces a vulnerability to error propagation. An incorrect fact extraction or flawed logical inference in an early step can cascade, adversely affecting all subsequent steps and the final outcome. This lack of fault tolerance presents a challenge for the deployment of fully autonomous reasoning systems. Future research should focus on integrating robustness mechanisms such as self-consistency or verification prompting.

The MoBLLM model contributes to precise and robust provision of personalized information for transportation management, e.g., crowding and disruption management. However, note that the incident logs or event information are important for prediction. Future work will involve the design of context-aware prompting strategies and the exploration of fine-tuning techniques to enhance MoBLLM’s adaptability, particularly in abnormal situations. Another promising direction involves enhancing the interpretability of LLM-generated predictions to gain deeper insights into the mechanisms governing human mobility behaviors. We also plan to extend and redevelop the proposed foundational model by evaluating its transferability to a wider spectrum of mobility prediction tasks, such as trip duration estimation. Additionally, we will explore practical applications of MoBLLM in personalized mobility research, including trip pattern analysis, to demonstrate its broader utility.

\section{CRediT authorship contribution statement}
\textbf{Zhenlin Qin}: Conceptualization, Data curation, Methodology, Visualization, Investigation, Formal analysis, Validation, Writing – original draft.
\textbf{Leizhen Wang}: Methodology, Validation,  Writing – review and editing.
\textbf{Yangcheng Ling}: Data curation, Validation
\textbf{Zhenliang Ma}: Conceptualization, Data curation, Methodology, Formal 
analysis, Writing – review and editing, Supervision.
\textbf{Francisco Câmara Pereira}: Conceptualization, Formal 
analysis, Writing – review and editing, Supervision

\section*{Acknowledgements}
The work was supported by the TRENoP center (Swedish Strategic Research Area in Transportation) at KTH Royal Institute of Technology, Sweden.

\section{Data availability}
The code, datasets (upon confidential check), experiments for this paper will be made public available at: https://github.com/qzl408011458/MoBLLM.



\appendix
\section{Instruction Examples}
\label{apx_instru_eg}
The following four instruction examples are respectively generated based on the instruction templates of corresponding base tasks defined in Section \ref{sec_task}.

\setcounter{figure}{0}
\renewcommand{\thefigure}{\Alph{section}\arabic{figure}} 

\begin{figure}[ht]
\centerline{\includegraphics[width=0.8\textwidth]{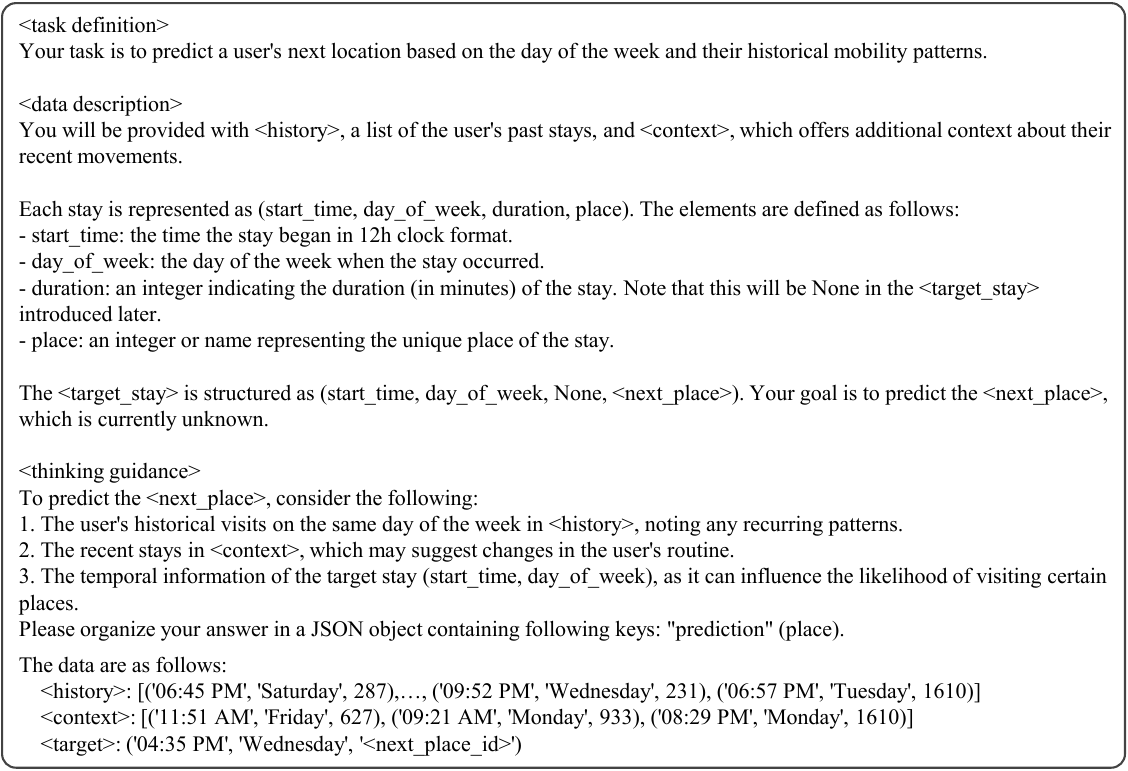}}
\caption{The generated instruction example for the task 1.}
\end{figure}

\begin{figure}[ht]
\centerline{\includegraphics[width=0.8\textwidth]{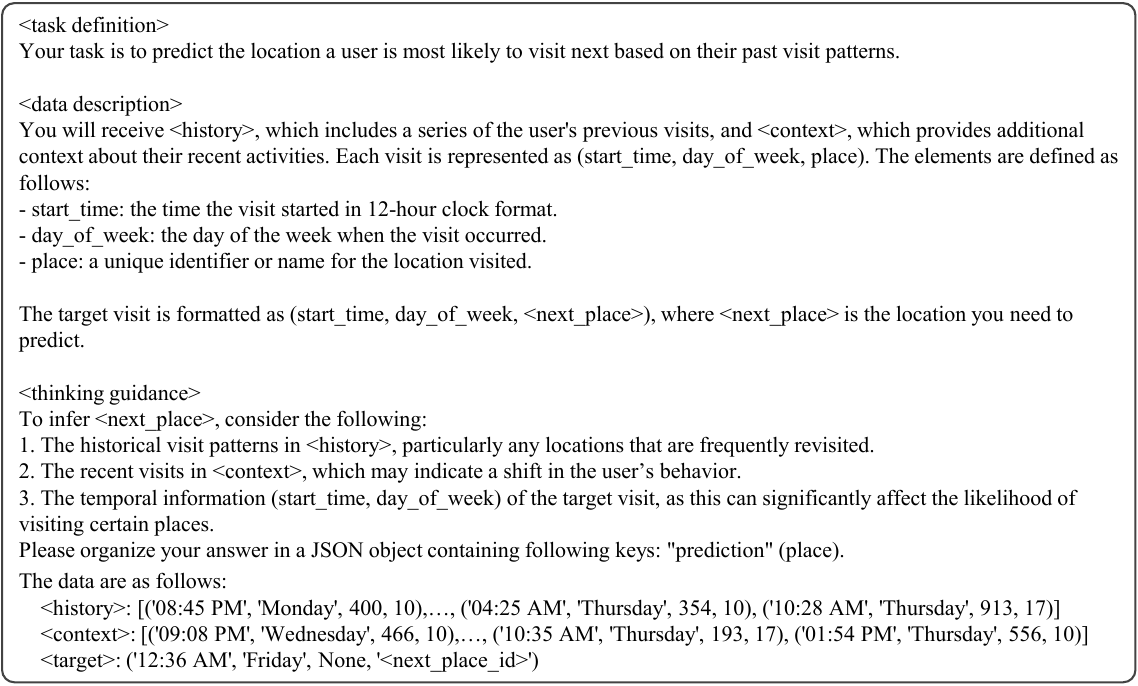}}
\caption{The generated instruction example for the task 2.}
\end{figure}

\begin{figure}[ht]
\centerline{\includegraphics[width=0.8\textwidth]{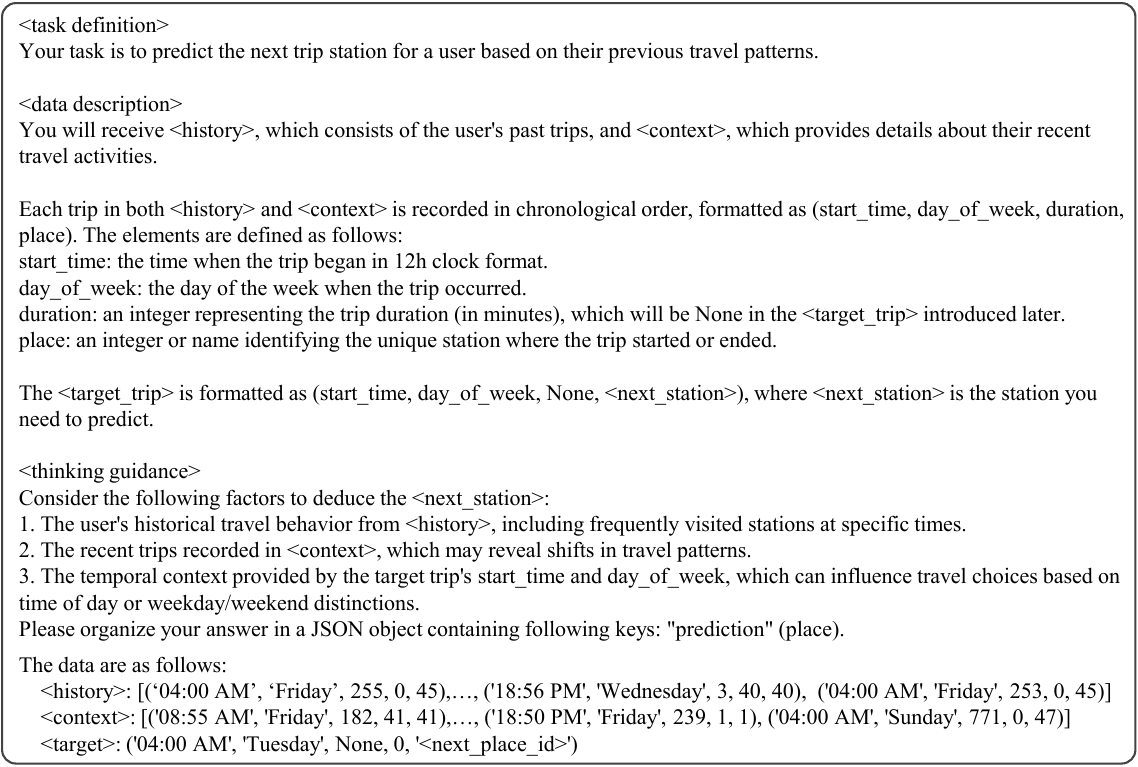}}
\caption{The generated instruction example for the task 3.}
\end{figure}

\begin{figure}[ht]
\centerline{\includegraphics[width=0.8\textwidth]{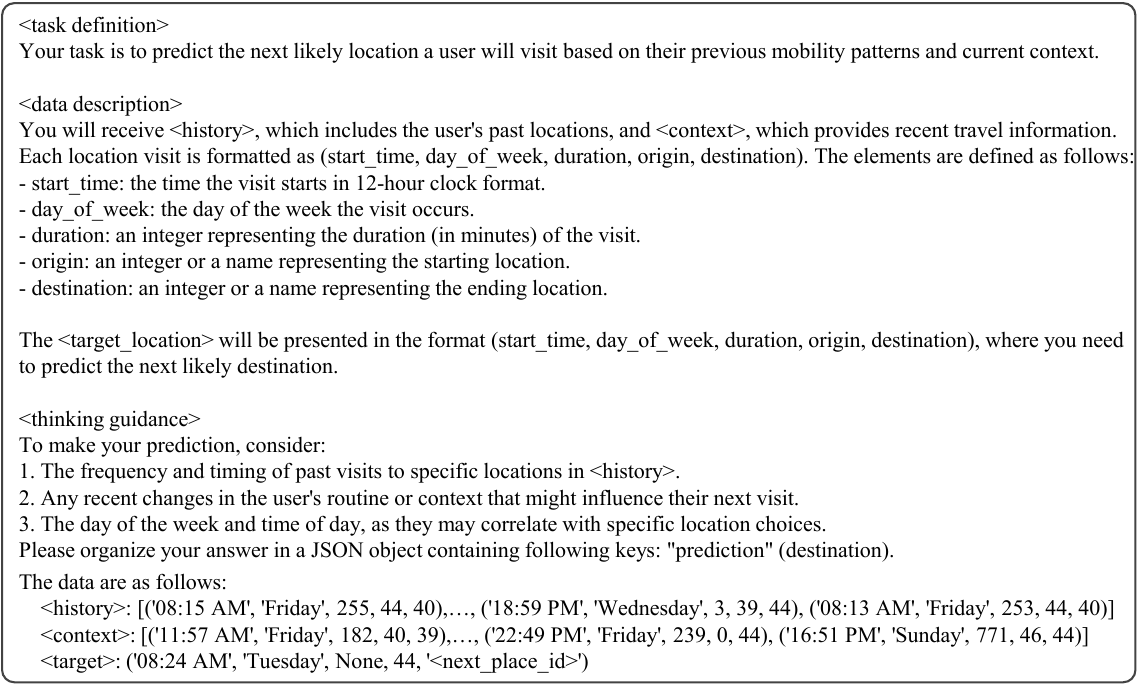}}
\caption{The generated instruction example for the task 4.}
\end{figure}

\FloatBarrier

\section{Hallucination Cases}
For the base LLaMA-3.1-8B-Instruct model, we test it on the datasets used in the case studies using the same prompts as the MoBLLM model. As shown in Figures \ref{fig_hall1} and\ref{fig_hall2}, the base model often produces irrelevant content, such as Python code snippets unrelated to the prediction task, which we identify as hallucinations in this context.

\setcounter{figure}{0}
\renewcommand{\thefigure}{\Alph{section}\arabic{figure}} 

\begin{figure}[ht]
\centerline{\includegraphics[width=0.8\textwidth]{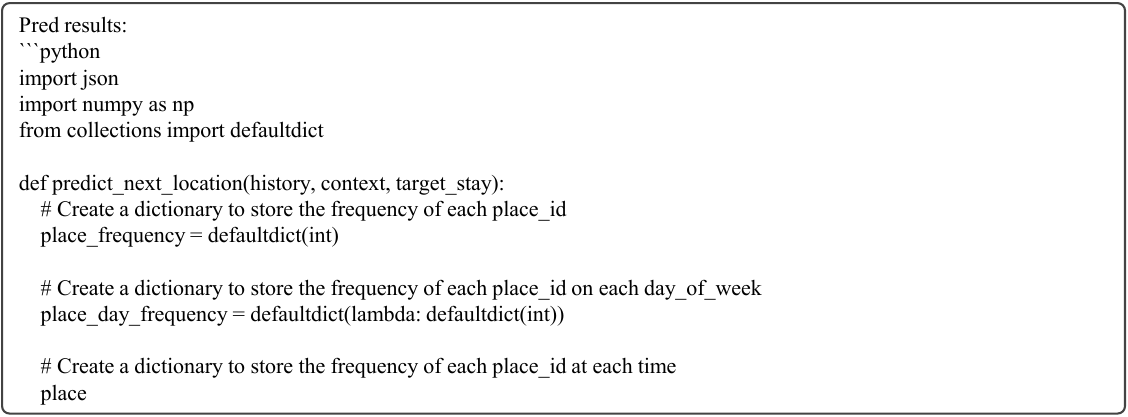}}
\caption{The hallucination example for the task 1.}
\label{fig_hall1}
\end{figure}

\begin{figure}[ht]
\centerline{\includegraphics[width=0.8\textwidth]{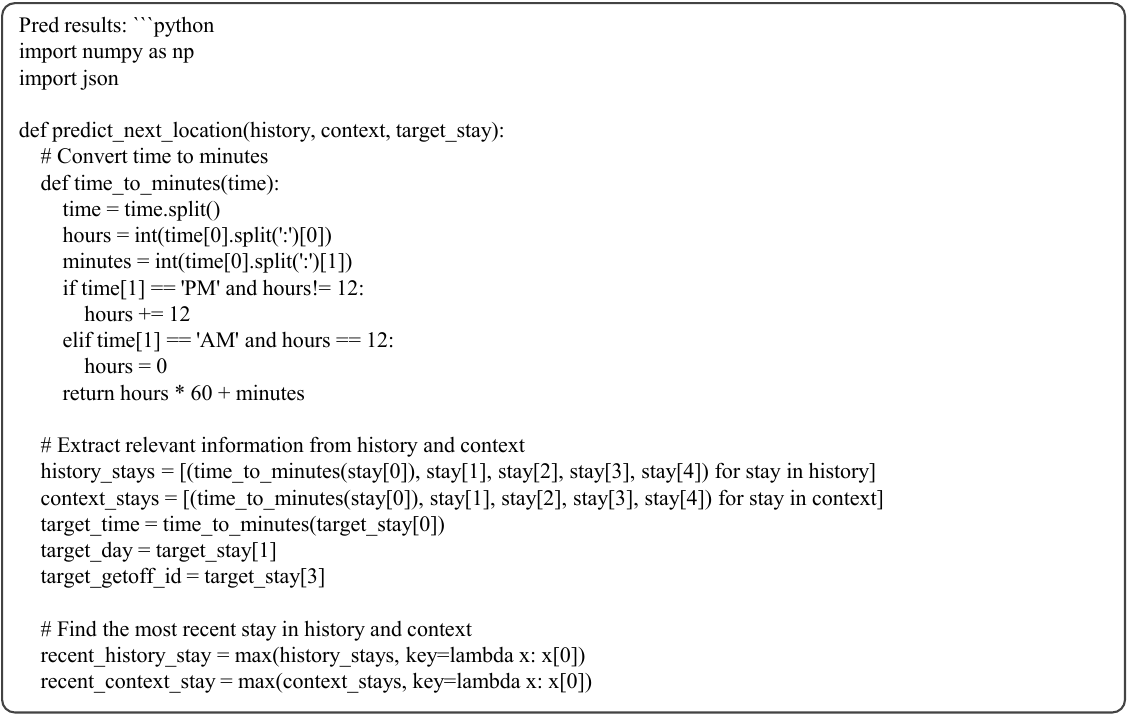}}
\caption{The hallucination example for the task 3.}
\label{fig_hall2}
\end{figure}

\FloatBarrier

\section{Hyperparameter Settings of PEFT}
For PEFT methods tested in Section \ref{sec_peft}, we present the settings of essential hyperparameter scaling factor $\alpha$ and rank $r$ as shown in Table \ref{lora_hyper}.

\setcounter{table}{0}
\renewcommand{\thefigure}{\Alph{section}\arabic{table}} 

\begin{table}[]
\centering
\caption{The hyperparameter settings of PEFT methods.}
\begin{tabular}{lll}
\hline
\textbf{PEFT} & \textbf{Scaling Factor} & \textbf{Rank} \\ \hline
LoRA          & 64                      & 64            \\
OLoRA         & 16                      & 64            \\
EVA           & 1                       & 16            \\
PiSSA         & 64                      & 64            \\
LoftQ         & 16                      & 64            \\
LoRA+         & 8                       & 32            \\
rsLoRA        & 8                       & 32            \\
QLoRA         & 8                       & 32            \\ \hline
\end{tabular}
\label{lora_hyper}
\end{table}

\section{Ablation Study on Instruction Training Data}
In this study, we fine-tune the LLaMA-3.1-8B-Instruct model using single-style instruction data, while keeping all other experimental conditions consistent with those of MoBLLM. We referred to this variant as MoBLLM-S. The training set incorporates the same mobility data as MoBLLM but is constructed exclusively with predefined base task prompt templates, without introducing stylistic diversity. 

To assess the model’s robustness to diverse user writing styles, input prompts are generated following an approach analogous to that outlined in Section \ref{inst_data_gen}. Unlike the single-style input prompts aligned with LLM-Mob used in Section \ref{sec4}, the prompts in this experiment are generated by DeepSeek-Chat with a temperature of 0.3 based on the base task prompt templates to simulate a variety of writing styles. Note that employing a different LLM (DeepSeek-Chat) for generating test cases prevents leakage of styles identical to those produced by GPT-4o mini (generating training instructions for MoBLLM), thereby ensuring a fair comparison. 

The results of MoBLLM-S and MoBLLM are presented in Table \ref{tab_ab}, which demonstrate the advantages of multi-style instructions for fine-tuning to achieve higher model performance.

\setcounter{table}{0}
\renewcommand{\thefigure}{\Alph{section}\arabic{table}} 

\begin{table}[]
\centering
\caption{Ablation study on instruction training data}
\begin{tabular}{lllllllll}
\hline
\multirow{2}{*}{\textbf{Model}} & \multicolumn{2}{l}{Geolife} & \multicolumn{2}{l}{FSQ-NYC} & \multicolumn{2}{l}{HK-ORI} & \multicolumn{2}{l}{HK-DEST} \\ \cline{2-9} 
                                & ACC          & F1           & ACC          & F1           & ACC         & F1           & ACC          & F1           \\ \hline
MoBLLM-S                        & 57.79        & 0.5318       & 40.98        & 0.3630       & 83.71       & 0.8362       & 70.46        & 0.6991       \\
MoBLLM                          & 58.37        & 0.5381       & 41.14        & 0.3638       & 84.26       & 0.8421       & 71.44        & 0.7093       \\ \hline
\end{tabular}
\label{tab_ab}
\end{table}

\section{Model Performance for In-/Frequent Visited Locations}
This study further investigates model performance on visited locations with different frequencies. First, we define the visited frequency as the ratio of the target location (ground truth) appearing in the model input (mobility sequence). To better present the stratified results, we select Geolife, FSQ-NYC and FSQ-TYK datasets and split their samples into 4-level visited frequencies as shown in Table \ref{tab_sample_split}. The results of Table \ref{tab_freq_comp1} show that both LLM-Mob and MoBLLM models perform well in high-frequency samples but hardly make predictions on the target locations that have never been seen in the model input. Compared with the commercial LLM for general purpose, MoBLLM performs significantly better in the low- to medium- frequency range from 0 to 0.5, implying that fine-tuning can effectively improve the LLMs' performance in predicting samples with such visited frequencies.

\setcounter{table}{0}
\renewcommand{\thefigure}{\Alph{section}\arabic{table}} 

\begin{table}[]
\centering
\caption{The sample proportions (\%) of 4-level visited frequencies.}
\begin{tabular}{llll}
\hline
\textbf{Frequency} & \textbf{Geolife} & \textbf{FSQ-NYC} & \textbf{FSQ-TYK} \\ \hline
0                  & 12.44            & 1.61             & 1.07             \\
(0, 0.2{]}         & 32.58            & 66.77            & 86.89            \\
(0.2, 0.5{]}       & 44.72            & 27.22            & 11.52            \\
(0.5, 1{]}             & 10.26            & 4.4              & 0.52             \\ \hline
\end{tabular}

\label{tab_sample_split}
\end{table}

\begin{table}[]
\centering
\caption{The performance comparison (ACC) under different visited frequencies.}
\begin{tabular}{lllllll}
\hline
\multicolumn{1}{c}{\multirow{2}{*}{\textbf{Frequency}}} & \multicolumn{2}{c}{\textbf{Geolife}} & \multicolumn{2}{c}{\textbf{FSQ-NYC}} & \multicolumn{2}{c}{\textbf{FSQ-TYK}} \\ \cline{2-7} 
\multicolumn{1}{c}{}                                    & LLM-Mob           & MoBLLM           & LLM-Mob           & MoBLLM           & LLM-Mob           & MoBLLM           \\ \hline
0                                                       & 0                 & 0.93             & 0                 & 0                & 0                 & 0                \\
(0, 0.2{]}                                              & 29.9              & 44.63            & 12.06             & 25.27            & 10.24             & 22.09            \\
(0.2, 0.5{]}                                            & 66.77             & 77.38            & 61.27             & 73.68            & 59.03             & 78.16            \\
(0.5, 1{]}                                                   & 90.14             & 87.61            & 90.24             & 92.68            & 94.92             & 96.61            \\ \hline
\end{tabular}
\label{tab_freq_comp1}
\end{table}

\bibliographystyle{elsarticle-harv} 
\bibliography{_mybib}

\end{document}